\documentclass[journal]{IEEEtran}

\usepackage{cite}
\usepackage{amsmath,amssymb,amsfonts}
\usepackage{algorithmic}
\usepackage{graphicx}
\usepackage{textcomp}
\usepackage{booktabs}
\usepackage{multirow}
\usepackage[ruled,linesnumbered]{algorithm2e}
\usepackage[hidelinks]{hyperref}

\begin{document}

\title{Bound-Constrained Sparse Representation for Electrical Impedance Tomography}

\author{Chun Zhang and Dong Liu, \IEEEmembership{Senior Member, IEEE}
\thanks{This work was supported by the National Natural Science Foundation of China under Grant 62371433.}
\thanks{C. Zhang and D. Liu are with the School of Biomedical Engineering and Suzhou Institute for Advanced Research, University of Science and Technology of China, Suzhou 215123, China. Corresponding author: Dong Liu (e-mail: dong.liu@outlook.com).}
\thanks{D. Liu is also with the Laboratory of Spin Magnetic Resonance and Anhui Province Key Laboratory of Scientific Instrument Development and Application, University of Science and Technology of China, Hefei 230026, China, and with the Jiangsu Provincial Key Laboratory of Multimodal Digital Twin Technology, Suzhou 215123, China.}
\thanks{D. Liu is also with the Institute of Quantum Sensing of WuXi, Wuxi 214100, China.}
}

\maketitle

\begin{abstract}
This study proposes a bound-constrained sparse representation (BC-SR) framework for electrical impedance tomography (EIT), aimed at improving conductivity estimation without explicit regularization. BC-SR adopts a representation-driven strategy, generating conductivity from low-dimensional latent variables via an implicit composite parameterization. Structural priors are embedded using a truncated graph-Laplacian basis, while a bound-preserving nonlinear mapping enforces admissible conductivity ranges and improves conditioning through implicit gradient modulation. The approach ensures robust convergence, even under noisy or incomplete data. Extensive validation on 2D/3D simulations, tank experiments, and in-vivo lung data shows that BC-SR improves physical consistency and structural fidelity, offering enhanced robustness compared to traditional methods. Additionally, BC-SR enables 3D time-difference EIT reconstruction, offering improved spatial resolution and a more coherent representation of 3D conductivity distributions, particularly for in-vivo lung data. This suggests potential for improved performance in EIT, particularly in clinical applications for respiratory monitoring.
\end{abstract}


\section{Introduction}
\label{sec:introduction}

\IEEEPARstart{E}{lectrical} impedance tomography (EIT) is a non-invasive imaging modality with advantages including the absence of ionizing radiation, low cost, and high temporal resolution, making it suitable for real-time bedside monitoring in applications such as lung and cardiac function assessment \cite{qu2025early, brabant2022tidalvolume, bachmann2018electrical, cao2020real}. Despite these advantages, obtaining accurate and reliable reconstructions in practical settings remains challenging.

This challenge arises from two fundamental aspects: the highly nonlinear relationship between boundary measurements and internal conductivity, and the severe ill-posedness of the inverse problem \cite{2017direct_partial_data}. As a result, the reconstruction is highly sensitive to measurement noise and modeling errors \cite{brazey2022robust}.

To alleviate this difficulty, linearized approaches are commonly adopted, where measurements are interpreted relative to a reference state and the forward model is expanded around a baseline conductivity, yielding a stabilized linear inverse problem. This formulation underlies classical methods such as NOSER \cite{cheney1990noser} and GREIT \cite{2D_GREIT, 3D_GREIT}. Owing to their robustness to modeling uncertainties, such methods are widely used in differential imaging formulations.

In contrast, when no reference state is available, absolute EIT aims to reconstruct conductivity directly from a single set of measurements by solving a nonlinear data-misfit minimization problem. The resulting ill-posedness is commonly addressed through explicit regularization, including total variation, Laplacian smoothing, and Bayesian priors \cite{borsic2009vivo, gonzalez2017isotropic, wu2022bayesian_LapPrior}, with further improvements achieved via structured sparsity and hierarchical Bayesian modeling \cite{gong2017eit, liu2018baysLearning}. While effective, these formulations rely on externally imposed priors, leading to a trade-off between data fidelity and prior assumptions and requiring careful tuning of regularization parameters.

These limitations motivate alternative formulations in which prior knowledge is incorporated directly into the conductivity representation, rather than enforced through explicit regularization. From this perspective, representation-driven methods constrain the solution space to a structured low-dimensional manifold. Early approaches include analytic transforms such as DCT/DWT and orthogonal polynomial bases, as well as learned dictionaries (e.g., K-SVD) \cite{DCT_EIT, see2007image, zhang2019image, wang2022research}, which provide compact representations but remain fundamentally linear and thus limited in capturing complex conductivity variations.

To improve expressiveness, explicit or implicit representation strategies have been well explored, extending classical simple parameterizations to more flexible and structure-aware representations. A representative class includes shape-based methods \cite{liu2020shapeB2, ren_shape, liuGS_levelset}, which describe conductivity through implicit geometric evolution and are particularly suited for piecewise-constant or boundary-dominated structures. Another line of work employs localized basis expansions, such as Gaussian-type kernels \cite{liu2025gsr}, which provide spatial adaptivity while enforcing smoothness and locality via explicit basis construction.

Moving beyond explicitly constructed representations, more general parameterizations have been developed based on learning-based models. A representative class includes variational autoencoders (VAEs) \cite{2019learning_vae, 2022vaeEIT}, which learn low-dimensional latent representations that capture high-level features of EIT images, enabling compact and data-driven parameterizations of conductivity. More broadly, conductivity can be modeled through implicit neural representations, where Deep image prior (DIP)-style frameworks and coordinate-based networks with positional encoding \cite{liu2023deepeit, wang2023unsupervised, chuyu2025pami} parameterize conductivity via network weights and architectural bias, allowing regularization of the inverse problem without explicit training data. Physics-informed neural networks (PINNs) \cite{ICLR_PINNs, PGCT-PINN, SMYL_PINNs} further incorporate governing equations through PDE residual minimization to enforce physical consistency with the forward model. Despite their enhanced representational capacity, these approaches often rely on complex network architectures, resulting in high computational cost and sensitivity to initialization. In addition, when strong data-driven priors are imposed, the reconstruction may be dominated by the learned representation, effectively reducing the influence of the data fidelity term and limiting robustness under modeling uncertainties.

To address these limitations, we propose a bound-constrained sparse representation (BC-SR) framework for EIT, which introduces a composite nonlinear parameterization that generates conductivity from low-dimensional latent variables. Structural priors are incorporated via a truncated graph-Laplacian basis, while admissible conductivity ranges are enforced through an implicit nonlinear mapping. This formulation restricts the solution space to physically plausible configurations and enables efficient latent-space optimization using a damped Gauss--Newton scheme, providing a unified framework for both absolute and time-difference imaging.

The main contributions of this work are summarized as follows:

\begin{itemize}

    \item {\bf BC-SR Framework}: We propose a bound-constrained sparse representation framework that integrates structural priors and physical feasibility into a unified nonlinear parameterization of conductivity.

    \item {\bf Unified Representation across Imaging Regimes}: We develop a single conductivity representation that is applicable to both absolute and time-difference EIT, enabling consistent reconstruction under modeling uncertainties within a unified optimization framework.

    \item {\bf Comprehensive Validation}: The proposed method is validated on 2D/3D simulations, tank experiments, and in-vivo data, demonstrating improved structural fidelity, physical consistency, and robustness compared with conventional approaches.

\end{itemize}

The remainder of this paper is organized as follows. Section~\ref{sec:eit_forward} reviews the forward model; Section~\ref{pgsr_method} presents the proposed method; Section~\ref{sec:exp_setup} describes the experimental setup; Sections~\ref{sec:results_2d} and~\ref{sec:results_3d} report numerical and experimental results; Sections~\ref{discussion} and~\ref{conclusion} conclude the paper, with the former focusing on analysis and limitations, and the latter summarizing the work and outlining potential extensions.

\section{Forward Model of Electrical Impedance Tomography}
\label{sec:eit_forward}

EIT aims to recover the internal conductivity distribution $\kappa(\boldsymbol{x})$ inside a domain $\Omega$ from boundary voltage measurements induced by applied currents. Let $L$ electrodes be attached to $\partial\Omega$, and the governing equations are described by the complete electrode model (CEM) \cite{cheng1989electrode}:

\begin{align}
\nabla\!\cdot\!\big(\kappa(\boldsymbol{x})\nabla \phi(\boldsymbol{x})\big) &= 0,
&& \boldsymbol{x}\in\Omega, \nonumber\\
\phi + z_q\,\kappa\frac{\partial \phi}{\partial\boldsymbol{\nu}} &= U_q,
&& \boldsymbol{x}\in e_q, \ q=1,\ldots,L, \nonumber\\
\int_{e_q} \kappa\frac{\partial \phi}{\partial\boldsymbol{\nu}}\,\mathrm{d}S &= I_q,
&& q=1,\ldots,L, \nonumber\\
\kappa\frac{\partial \phi}{\partial\boldsymbol{\nu}} &= 0,
&& \boldsymbol{x}\in \partial\Omega\setminus\bigcup_{q=1}^L e_q.
\label{eq.cem}
\end{align}
where $\phi$ denotes the electric potential, $U_q$ and $I_q$ are electrode voltages and currents, and $z_q$ is the contact impedance. The standard charge conservation and reference potential constraints are imposed:
\[
\sum_{q=1}^{L} I_q = 0, \quad \sum_{q=1}^{L} U_q = 0.
\]

In the finite element discretization \cite{vauhkonen2002three}, the conductivity field is represented by a nodal vector $\boldsymbol{\sigma} \in \mathbb{R}^N$, where $N$ is the number of nodes in the FEM mesh.

The discrete forward model is given by
\begin{equation}
\boldsymbol{V} = \mathcal{U}(\boldsymbol{\sigma}) + \boldsymbol{e},
\label{eq.forward}
\end{equation}
where $\boldsymbol{V}$ denotes the measured electrode voltages and $\boldsymbol{e}$ is measurement noise.

\section{The proposed BC-SR Framework}
\label{pgsr_method}

\subsection{Inverse Problem Formulation}

Given boundary measurements $\boldsymbol{V}$ and the discrete forward mapping $\mathcal{U}(\cdot)$, EIT reconstruction aims to estimate the conductivity field $\boldsymbol{\sigma}$ by solving a bound-constrained nonlinear least-squares problem
\begin{equation}
\hat{\boldsymbol{\sigma}}
= \arg\min_{\boldsymbol{\sigma}}
\; \frac{1}{2}\big\|\mathcal{U}(\boldsymbol{\sigma}) - \boldsymbol{V}\big\|_2^2
\quad \text{s.t.} \quad
\boldsymbol{\ell} \le \boldsymbol{\sigma} \le \boldsymbol{u}.
\label{eq.inverse_formulation}
\end{equation}
Here, $\boldsymbol{\ell}$ and $\boldsymbol{u}$ denote element-wise lower and upper bound vectors, respectively. In this work, we assume a spatially uniform admissible conductivity range $[\ell, u]$ over the entire domain, such that $\boldsymbol{\ell} = (\ell,\ldots,\ell)^\top$ and $\boldsymbol{u} = (u,\ldots,u)^\top$.

Directly solving (\ref{eq.inverse_formulation}) requires handling bound constraints explicitly during optimization, which can complicate the solution process. To address this, we adopt an implicit nonlinear parameterization that encodes the admissible conductivity range directly into the model. Specifically, the conductivity is represented as
\begin{equation}
\boldsymbol{\sigma}=\mathcal{H}(\boldsymbol{c}), \qquad 
\boldsymbol{c}=\mathbf{B}\boldsymbol{\alpha},
\label{eq.pgsr_rep}
\end{equation}
where $\mathcal{H}(\cdot)$ is a nonlinear mapping that enforces the prescribed bounds, $\mathbf{B}$ is a fixed structural basis, and $\boldsymbol{\alpha}$ is a reduced coefficient vector.

With this parameterization, the bound constraints are satisfied by construction, and the original constrained problem is transformed into an unconstrained optimization problem in the reduced latent space:
\begin{equation}
\hat{\boldsymbol{\alpha}}
= \arg\min_{\boldsymbol{\alpha}}
\; \frac{1}{2}\big\|
\mathcal{U}\!\big(\mathcal{H}(\mathbf{B}\boldsymbol{\alpha})\big)
- \boldsymbol{V}
\big\|_2^2 .
\label{eq:alpha_opt}
\end{equation}
The conductivity field is then recovered as $\hat{\boldsymbol{\sigma}}=\mathcal{H}(\mathbf{B}\hat{\boldsymbol{\alpha}})$.
This formulation embeds physical feasibility directly into the parameterization while reducing the dimensionality of the optimization problem. In the following, we introduce the nonlinear mapping $\mathcal{H}(\cdot)$ that enforces the admissible range, followed by the construction of the structural basis $\mathbf{B}$ for low-dimensional representation.

\subsection{Nonlinear Bound Mapping}
\label{sec:Nonlinear Bound Mapping}

The nonlinear mapping $\mathcal{H}(\cdot)$ enforces the admissible
conductivity range in a physically consistent manner. Specifically,
\begin{subequations}\label{eq:sigmoid_shift}
\begin{align}
\mathcal{H}(\boldsymbol{c})
&=
\boldsymbol{\ell}
+
(\boldsymbol{u}-\boldsymbol{\ell})\odot
f(\boldsymbol{c}+\Delta\boldsymbol{c}),\\
f(z)
&=\frac{1}{1+e^{-z}},
\end{align}
\end{subequations}
where $\odot$ denotes element-wise multiplication.
By construction, $\mathcal{H}(\boldsymbol{c}) \in [\boldsymbol{\ell}, \boldsymbol{u}]$.

To ensure consistency with a prescribed initial conductivity
$\boldsymbol{\sigma}^0 \in [\boldsymbol{\ell}, \boldsymbol{u}]$,
the shift $\Delta\boldsymbol{c}$ is chosen such that
\begin{equation}
\mathcal{H}(\boldsymbol{0})=\boldsymbol{\sigma}^{0},
\end{equation}
which yields
\begin{equation}
\Delta\boldsymbol{c}
=
\log\!\left(
\frac{\boldsymbol{\sigma}^{0}-\boldsymbol{\ell}}
     {\boldsymbol{u}-\boldsymbol{\sigma}^{0}}
\right),
\end{equation}
with all operations applied componentwise.

In the absence of prior information, we adopt a homogeneous initialization
$\boldsymbol{\sigma}_\mathrm{homo}$ and estimate a global scaling factor
\begin{equation}
s = \frac{\sum_{i} V_i^2}
         {\sum_{i} [\mathcal{U}(\boldsymbol{\sigma}_\mathrm{homo})]_i \, V_i},
\label{eq:s_sigma0}
\end{equation}
so that $\boldsymbol{\sigma}^0 = s\,\boldsymbol{\sigma}_\mathrm{homo}$.

The mapping $\mathcal{H}$ also modulates the update sensitivity. Its derivative is given by
\begin{equation}
\mathcal{H}'(\boldsymbol{c}) =
\mathrm{diag}\!\left(
\frac{(\boldsymbol{\sigma} - \boldsymbol{\ell}) \odot (\boldsymbol{u} - \boldsymbol{\sigma})}{\boldsymbol{u} - \boldsymbol{\ell}}
\right),
\end{equation}
which naturally damps updates near the bounds and improves numerical stability.

Importantly, this formulation provides a unified framework for both absolute and time-difference imaging. 
In the difference imaging setting, let $\boldsymbol{\sigma}_0$ and $\boldsymbol{\sigma}_1$ denote the baseline and target states. By selecting $\Delta\boldsymbol{c}$ such that $\mathcal{H}(\boldsymbol{0}) = \boldsymbol{\sigma}_0$, the reconstruction is naturally warm-started as
\begin{equation}
\boldsymbol{\sigma}_1^{(0)} = \boldsymbol{\sigma}_0.
\end{equation}
Under this formulation, the reconstruction updates deviations from a physically consistent baseline, thereby reducing the impact of modeling errors and enabling robust time-difference imaging within the same optimization framework.

\subsection{Graph Laplacian-based Sparse Representation}
\label{sec:graph_laplacian_sparse_representation}

Within the BC-SR framework, the linear basis $\mathbf{B}$ encodes the structural prior of the conductivity field. On an FEM mesh, a natural choice is to exploit node connectivity. We construct an undirected graph from the mesh nodes, with adjacency matrix $\mathbf{W}$ defined by
\begin{equation}
w_{ij} =
\begin{cases}
1, & \text{if nodes $i$ and $j$ are connected},\\
0, & \text{otherwise}.
\end{cases}
\end{equation}
This unweighted connectivity-based construction avoids introducing additional tuning parameters while preserving the mesh topology.
The corresponding graph Laplacian is
\begin{equation}
\mathbf{L} = \mathbf{D} - \mathbf{W}, \qquad
\mathbf{D} = \mathrm{diag}\Big(\sum_j w_{ij}\Big),
\end{equation}
where the sum is taken over all nodes connected to node $i$.

The Laplacian matrix $\mathbf{L}$ is symmetric positive semidefinite and admits the eigen-decomposition
\begin{equation}
\mathbf{L} = \mathbf{G}\mathbf{\Lambda}\mathbf{G}^\top,
\end{equation}
where the columns of $\mathbf{G}$ are orthonormal eigenvectors, ordered according to increasing eigenvalues in $\mathbf{\Lambda}$.
In practice, these eigenvectors are obtained by solving the standard symmetric eigenvalue problem associated with $\mathbf{L}$. Since $\mathbf{L}$ is sparse and constructed purely from mesh connectivity,
efficient numerical solvers can be directly applied.
Eigenvectors corresponding to the smallest eigenvalues capture smooth, low-frequency modes on the mesh. We therefore define
\begin{equation}
\mathbf{B} := \mathbf{G}_{N_b},
\end{equation}
where  $\mathbf{G}_{N_b} \in \mathbb{R}^{N \times N_b}$ contains the first $N_b$ eigenvectors ($N_b \ll N$). This effectively performs a low-frequency dimensionality reduction on the mesh, analogous to DCT or other standard transforms for regular grids.

\begin{figure}[!htbp]
\centering
\includegraphics[width=.8\columnwidth]{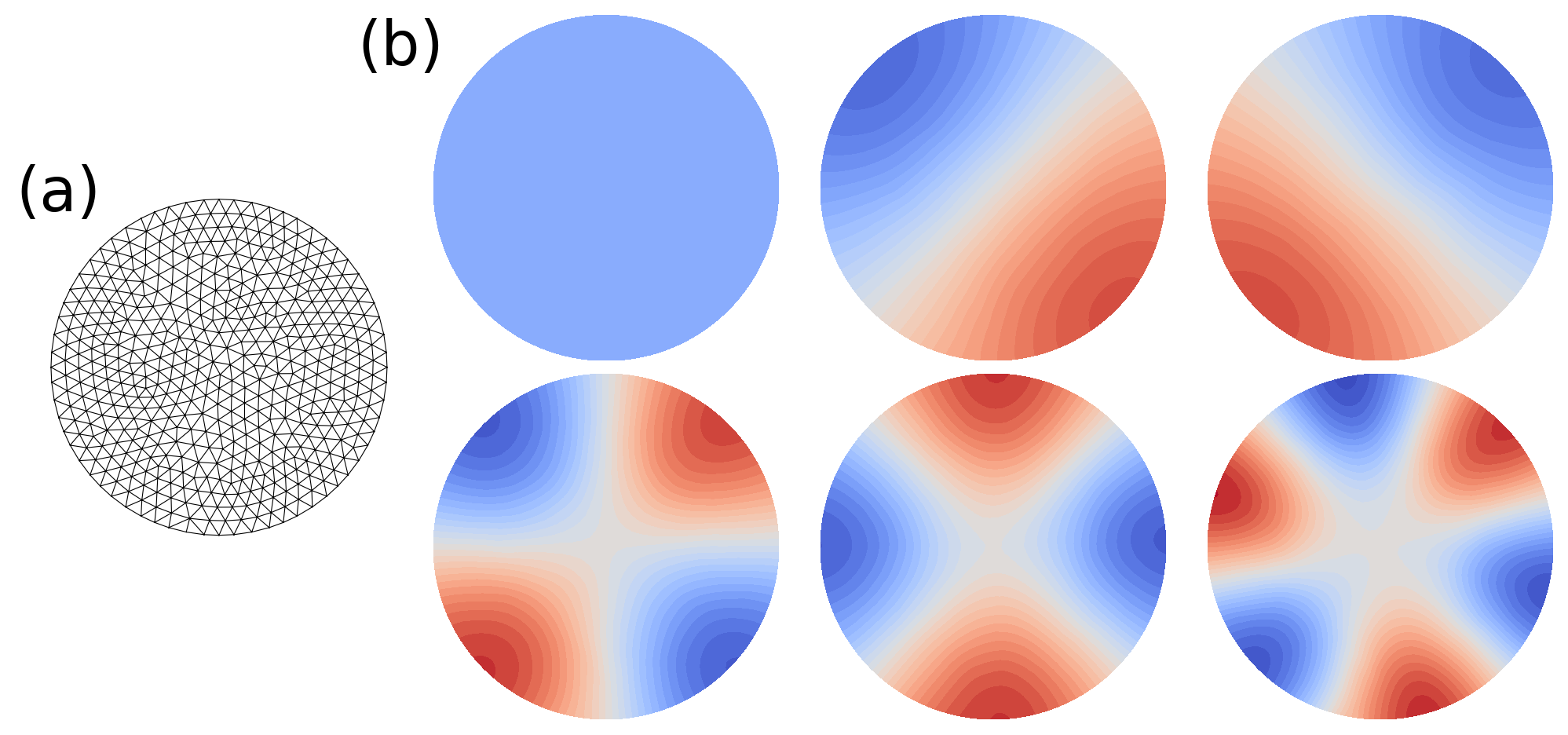}
\caption{
(a) Finite element mesh.  
(b) The first six Laplacian eigenvectors illustrating smooth, low-frequency
graph modes.}
\label{fig.lp_basis}
\end{figure}

Importantly, this construction is dimension-independent. For both 2D and 3D meshes, the graph Laplacian is assembled in the same manner from node connectivity, and the corresponding eigenvectors are computed using the same procedure. The only difference lies in the mesh structure itself, not in the formulation. Therefore, the proposed basis construction generalizes naturally to arbitrary geometries and unstructured meshes (see Fig.~\ref{fig.lp_basis}).

\subsection{Optimization Algorithm}
\label{sec:algorithm}

The BC-SR objective function defined in \eqref{eq:alpha_opt} is solved using the
Levenberg--Marquardt--Fletcher (LMF) algorithm \cite{fletcher2013practical},
a standard method for nonlinear least-squares optimization.

We define the residual as
\[
\boldsymbol{r}=\mathcal{U}(\boldsymbol{\sigma})-\boldsymbol{V},
\]
where $\boldsymbol{r} \in \mathbb{R}^{M}$,
$\boldsymbol{\sigma} \in \mathbb{R}^{N}$, and
$\boldsymbol{V} \in \mathbb{R}^{M}$.
By the chain rule, the Jacobian with respect to $\boldsymbol{\alpha}$ is given by
\begin{equation}
    \mathcal{J}_{\alpha}
    = \mathcal{J}_{\sigma}\, \mathcal{H}'(\boldsymbol{c})\, \mathbf{B},
\end{equation}
where $\mathcal{J}_{\sigma} = \frac{\partial \boldsymbol{r}}{\partial \boldsymbol{\sigma}} \in \mathbb{R}^{M \times N}$,
$\mathcal{H}'(\boldsymbol{c}) \in \mathbb{R}^{N \times N}$ is diagonal, and
$\mathbf{B} \in \mathbb{R}^{N \times N_b}$.
Accordingly, $\mathcal{J}_{\alpha} \in \mathbb{R}^{M \times N_b}$.

The gradient and Gauss--Newton approximation of the Hessian are
\begin{subequations}
\label{eq:LMF_grad_hess}
\begin{align}
    \nabla \mathcal{L}(\boldsymbol{\alpha})
        &= \mathcal{J}_{\alpha}^{\top}\boldsymbol{r}, \\
    \boldsymbol{H}(\boldsymbol{\alpha})
        &= \mathcal{J}_{\alpha}^{\top}\mathcal{J}_{\alpha} + \mu\mathbf{I},
\end{align}
\end{subequations}
where $\mu>0$ is a damping parameter, $\mathbf{I} \in \mathbb{R}^{N_b \times N_b}$ is the identity matrix.

At iteration $k$, the update direction $\boldsymbol{d}$ is obtained from
\begin{equation}
\label{eq:damped_normal_equation}
    \left(\mathcal{J}_{\alpha}^{\top}\mathcal{J}_{\alpha}
        + \mu_k\mathbf{I}\right)\boldsymbol{d}
    = -\mathcal{J}_{\alpha}^{\top}\boldsymbol{r}.
\end{equation}
The step is accepted based on the ratio of actual to predicted reduction, 
\begin{equation}
\rho_k = \frac{ \mathcal{L}(\boldsymbol{\alpha}^{k}) - \mathcal{L}(\boldsymbol{\alpha}^{k} + \boldsymbol{d}) }{ m_k(\mathbf{0}) - m_k(\boldsymbol{d}) }, 
\end{equation} 
and the damping parameter is updated following Fletcher’s rule 
\begin{equation}
\mu_{k+1}= \begin{cases} \gamma_2\mu_k, & \rho_k < \rho_1,\\ \gamma_1\mu_k, & \rho_k > \rho_2,\\ \mu_k, & \text{otherwise}. \end{cases} 
\label{eq:LMF_update} 
\end{equation}

\subsection{Implementation Details}
For clarity, we distinguish between the LMF-related hyperparameters
\mbox{$\{\mu_0,\gamma_1,\gamma_2,\rho_1,\rho_2\}$},
which govern the damping and trust-region updates, and the stopping parameters,
namely $T_{\max}$ and $\varepsilon$, which control the maximum number of iterations
and the minimum step size, respectively. 
Unless otherwise stated, we adopt standard textbook values for the LMF algorithm,
\[
\rho_{1}=0.25,\quad \rho_{2}=0.75,\quad
\gamma_{1}=0.25,\quad \gamma_{2}=4,
\]
and initialize the damping parameter with a small value $\mu_{0}$ (e.g.,
$\mu_{0}=10^{-4}$), which effectively starts the iterations in the
Gauss--Newton regime. These values are not tuned in our experiments and are kept identical across all 2D and 3D settings (Sections~\ref{subsec:setup_2d} and~\ref{subsec:setup_3d}), consistent with standard practice and the well-known robustness of the LMF algorithm.

The latent variables are initialized as
$\boldsymbol{\alpha}^{0}=\boldsymbol{0}$, which yields a homogeneous latent field
$\boldsymbol{c}^{0}=\boldsymbol{0}$ and a physically admissible background
conductivity $\boldsymbol{\sigma}^{0}$ defined in~\eqref{eq:s_sigma0}.
The outer iterations are terminated when either the maximum iteration count
$T_{\max}$ is reached or the step norm $\|\boldsymbol{d}^{k}\|$ falls below a
prescribed tolerance $\varepsilon$.
The complete BC-SR reconstruction procedure is summarized in
Algorithm~\ref{alg:PGSR}.

\begin{algorithm}[t]
\caption{BC-SR Reconstruction Algorithm}
\label{alg:PGSR}
\KwIn{Measurements $\boldsymbol{V}$, forward operator $\mathcal{U}$}
\KwIn{Graph basis $\mathbf{B}$, physical bounds $({\ell},{u})$}
\KwIn{Hyperparameters $\Theta = \{T_{\max},\varepsilon,\mu_0,
        \gamma_1,\gamma_2,\rho_1,\rho_2\}$}
\KwOut{Reconstructed conductivity $\hat{\boldsymbol{\sigma}}$}

\textbf{Initialization:}
$\boldsymbol{\alpha}^{0}=\boldsymbol{0}$;
$\boldsymbol{c}^{0}=\boldsymbol{0}$;
$\boldsymbol{\sigma}^{0}$ given by~\eqref{eq:s_sigma0};
$\mu_k \leftarrow \mu_0$;
$k\leftarrow 0$;

\BlankLine
\While{$k < T_{\max}$ \textbf{and} $\|\boldsymbol{d}^{k}\| > \varepsilon$}{
    Compute residual $\boldsymbol{r}^k=\mathcal{U}(\boldsymbol{\sigma}^k)-\boldsymbol{V}$\;
    Form latent Jacobian
    $\mathcal{J}_{\alpha}
        =\mathcal{J}_{\sigma}\,\mathcal{H}'(\boldsymbol{c}^k)\,\mathbf{B}$\;

    \Repeat{$\rho_k>\rho_1$}{
        Solve LMF step:
        $(\mathcal{J}_{\alpha}^{\top}\mathcal{J}_{\alpha}+\mu_k\mathbf{I})
        \boldsymbol{d} = -\mathcal{J}_{\alpha}^{\top}\boldsymbol{r}^k$\;

        $\hat{\boldsymbol{\alpha}}
            = \boldsymbol{\alpha}^{k}+\boldsymbol{d}$\;
        Compute reduction ratio $\rho_k$\;
        Update $\mu_k$ according to \eqref{eq:LMF_update}\;
    }

    Accept the step\;
    $\boldsymbol{\alpha}^{k+1}=\hat{\boldsymbol{\alpha}}$\;
    $\boldsymbol{c}^{k+1}=\mathbf{B}\boldsymbol{\alpha}^{k+1}$\;
    $\boldsymbol{\sigma}^{k+1}=\mathcal{H}(\boldsymbol{c}^{k+1})$\;

    $k\leftarrow k+1$\;
}
$\hat{\boldsymbol{\sigma}} \leftarrow \boldsymbol{\sigma}^{k}$\;
\Return{$\hat{\boldsymbol{\sigma}}$}
\end{algorithm}

\subsection{Computational Efficiency}

BC-SR achieves computational efficiency through two design choices.
First, the graph-Laplacian sparse representation reduces the optimization
dimension from $N$ to $N_{b}$ ($N_{b}\ll N$), thereby decreasing the cost of solving
the normal equations. The basis $\mathbf{B}$ is precomputed and fixed,
and the bound mapping $\mathcal{H}$ is applied componentwise.
The latent Jacobian
$\mathcal{J}_{\alpha}=\mathcal{J}_{\sigma}\,\mathcal{H}'(\boldsymbol{c})\,\mathbf{B}$
is assembled with minimal additional cost beyond $\mathcal{J}_{\sigma}$.

Second, BC-SR employs the LMF algorithm, requiring one solve of the damped
normal equations per iteration (see \eqref{eq:damped_normal_equation}).
In practice, convergence is achieved within approximately 10 iterations
on the in-vivo data considered in this study.

The overall runtime can be approximated as
\begin{equation}
    T \approx N_{\mathrm{iter}}\bigl(T_{\mathrm{fwd}} + T_{\mathrm{J}}\bigr),
\end{equation}
where $T_{\mathrm{fwd}}$ denotes forward/adjoint solves and
$T_{\mathrm{J}}$ the Jacobian assembly cost.



\section{Experimental Setup}
\label{sec:exp_setup}
\subsection{Measurement Protocol and Evaluation Metrics}
\label{subsec:measurement_protocol}
All experiments are conducted using a 16-electrode configuration. 
Adjacent current injection and adjacent voltage measurement patterns are adopted as the default protocol.

For numerical simulations (2D and 3D), all adjacent stimulation patterns are used together with adjacent voltage measurements.
For the tank experiments, a subset of stimulation patterns is used by selecting electrodes 
$\{1,5,9,13\}$ as one terminal and enumerating all distinct pairings, resulting in 54 independent stimulation patterns. 
Adjacent voltage measurements are employed.
For the in-vivo data, adjacent stimulation is applied, and voltage measurements involving current-driven electrodes are excluded. 
Reciprocal measurements are further removed, resulting in 104 independent measurements per frame.

Synthetic voltages $\boldsymbol{V}$ are generated using the forward model in \eqref{eq.forward}, with additive Gaussian noise corresponding to a signal-to-noise ratio (SNR) of 60~dB:
\begin{equation}
    \text{SNR} = 10 \log_{10}\frac{\sum_i \left| \mathcal{U}(\boldsymbol{\sigma})_i \right|^2}{\sum_i \left| \boldsymbol{e}_i \right|^2}.
\end{equation}

To avoid inverse crimes, different FEM meshes are used for forward simulation and reconstruction. 
All conductivity values are reported in mS/cm.
Reconstruction quality is evaluated using Structural Similarity Index (SSIM)~\cite{SSIM_definition}, correlation coefficient (CC), and root mean square error (RMSE).

\subsection{Reconstruction Methods}
\label{subsec:methods}
We compare the proposed BC-SR method with conventional voxel-wise reconstruction approaches, including NOSER~\cite{cheney1990noser}, iterative $\ell_2$~\cite{L2EIT}, and TV~\cite{borsic2009vivo}, which are widely used and serve as standard baselines.

As a reference method for time-difference imaging, we adopt a linearized difference (LD) approach. The forward model is linearized around a baseline conductivity $\boldsymbol{\sigma}_0$:
\begin{equation}
\boldsymbol{V}_i \approx \boldsymbol{V}_0 + \mathbf{J}_0\,\Delta\boldsymbol{\sigma}_i,
\end{equation}
where $\mathbf{J}_0$ denotes the Jacobian evaluated at $\boldsymbol{\sigma}_0$.

The reconstruction is obtained by solving the following regularized least-squares problem:
\begin{equation}
\widehat{\Delta\boldsymbol{\sigma}}_i
=
\arg\min_{\Delta\boldsymbol{\sigma}}
\left\|
\Delta \boldsymbol{V}_i - \mathbf{J}_0 \Delta \boldsymbol{\sigma}
\right\|_2^2
+
\alpha \Delta\boldsymbol{\sigma}^T \mathbf{R} \,\Delta\boldsymbol{\sigma},
\end{equation}
where $\mathbf{R} = \mathrm{diag}(\mathbf{J}_0^T \mathbf{J}_0)$ corresponds to a NOSER-type regularization.
This formulation admits a closed-form solution of the form
\begin{equation}
\widehat{\Delta\boldsymbol{\sigma}}_i
=
\left(\mathbf{J}_0^T \mathbf{J}_0 + \alpha \mathbf{R}\right)^{-1}
\mathbf{J}_0^T \Delta \boldsymbol{V}_i
=
\mathbf{A}\,\Delta \boldsymbol{V}_i,
\end{equation}
where $\mathbf{A}$ is the corresponding linear reconstruction operator. In this work, this formulation is consistently referred to as the LD baseline in all numerical experiments.

In contrast, BC-SR performs nonlinear reconstruction in a low-dimensional latent space. For time-difference imaging, a warm-start strategy is adopted by initializing the target state with the baseline solution, which improves robustness to modeling mismatch and facilitates convergence.
\subsection{2D Studies}
\label{subsec:setup_2d}

\textbf{Ablation Study:}
We conduct ablation experiments on a 2D phantom (Case~1; Fig.~\ref{fig:ablation_H}), consisting of three homogeneous regions with conductivities $0.25$, $1.0$, and $2.0~\mathrm{mS/cm}$. 

Two aspects are evaluated: (i) constraint enforcement via nonlinear bound mapping (NLM) versus an active-set method (ASM), under fine bounds $[0.2,2.0]$ and coarse bounds $[0.1,4.0]$, and (ii) the effect of the graph-Laplacian sparse representation (SR) and basis truncation.

\vspace{0.5em}
\noindent
\textbf{2D Numerical Phantoms:}
Four phantoms (Cases~2--5) are considered, covering smooth, sharp, high-contrast, and lung-like conductivity distributions. Two BC-SR variants are evaluated, which differ only in the conductivity bounds: BC-SR(F) uses accurate bounds consistent with the ground-truth range, while BC-SR(C) adopts relaxed bounds with a wider admissible interval.

\vspace{0.5em}
\noindent
\textbf{Noise Robustness Study:}
To improve stability under noise, a lightweight TV term is incorporated, with its weight automatically scaled by the SNR:
\begin{equation}
    \lambda(\mathrm{SNR})
    =
    \begin{cases}
        \lambda_0\left(\sqrt{10}\right)^{\frac{60 - \mathrm{SNR}}{10}}, & \mathrm{SNR} < 60~\mathrm{dB}, \\
        \lambda_0, & \mathrm{SNR} \ge 60~\mathrm{dB}.
    \end{cases}
    \label{eq:alpha_snr}
\end{equation}
\vspace{0.5em}
\noindent
\textbf{Tank Experiments:}
Experiments are conducted on a saline tank (28~cm diameter) with insulating inclusions. Five configurations (Tank~1--Tank~5) are evaluated, with conductivity bounds set to $[0.001,\,0.543]~\mathrm{mS/cm}$. Data are acquired using the KIT4 system~\cite{KIT4system}.

\subsection{3D Studies}
\label{subsec:setup_3d}

\begin{figure}[!htbp]
    \centering
    \includegraphics[width=.7\linewidth]{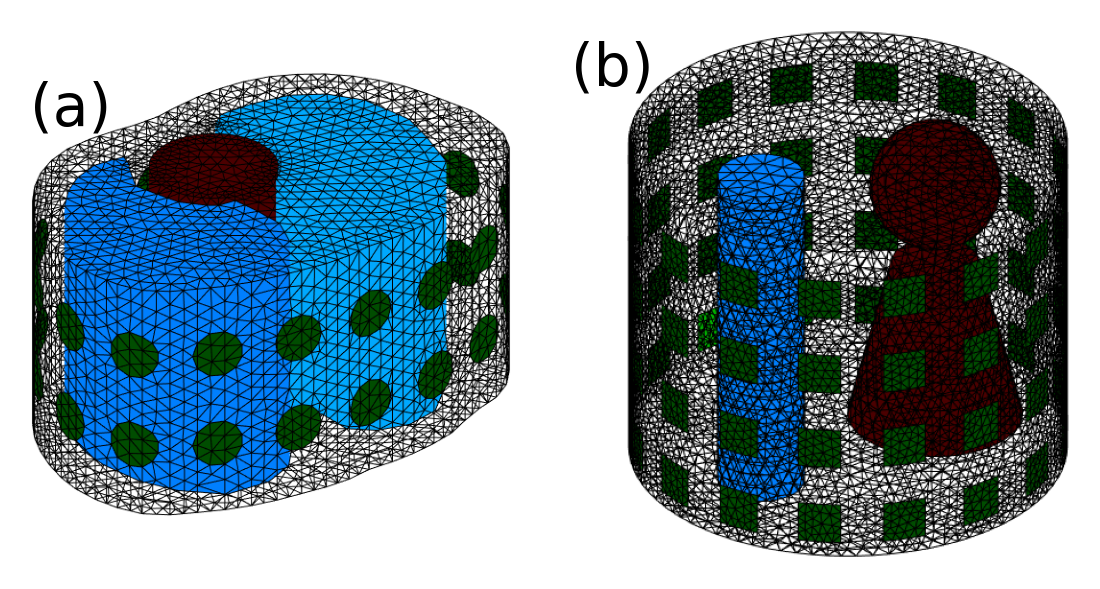}
    \caption{3D numerical phantoms and electrode configurations. Case~2.5D uses two electrode rings, and Case~3D uses four rings.}
    \label{fig:3d_phantoms}
\end{figure}

\textbf{3D Numerical Phantoms:}
We evaluate BC-SR on two 3D phantoms (2.5D and full 3D), with conductivity settings summarized in Table~\ref{tab:phantom_params} (see Fig.~\ref{fig:3d_phantoms}). Results are reported in a normalized coordinate system. Case~2.5D spans \(x \in [-0.945, 1.003]\), \(y \in [-0.651, 0.682]\), and \(z \in [-0.5, 0.5]\), while Case~3D uses \(x, y \in [-1, 1]\) and \(z \in [-0.8, 0.8]\).

\begin{table}[!htbp]
    \centering
    \caption{Conductivities of the 3D numerical phantoms (in mS/cm).}
    \label{tab:phantom_params}
    \setlength{\tabcolsep}{4pt}
    \begin{tabular}{lcccc}
        \toprule
        Case & Background & Region A & Region B & Region C  \\
        \midrule
        2.5D & 1.0 & 0.1 (left lung)  & 0.4 (right lung) & 3.0 (heart) \\
        3D & 1.0 & 0.5 (cylinder)   & 5.0 (sphere)     & 5.0 (frustum) \\
        \bottomrule
    \end{tabular}
\end{table}

\vspace{0.5em}
\noindent
\textbf{In-vivo Experiments:}
The data was obtained from a previously reported in-vivo EIT–CT study, where ethical approval and informed consent were obtained \cite{zhang2020supervised}. The normalized domain spans $x\in[-1,1]$, $y\in[-0.641,0.641]$, and $z\in[-0.667,0.667]$. Conductivity bounds are set to $[0.01,\,8]~\mathrm{mS/cm}$.

Time-difference imaging is performed over a respiratory cycle, with the end-expiration state ($t_0=40.2$~s) as the baseline. For each frame $t_i$, the conductivity change is defined as
\begin{equation}
\Delta \boldsymbol{\sigma}_i = \boldsymbol{\sigma}_i - \boldsymbol{\sigma}_0.
\end{equation}

\section{2D Results}
\label{sec:results_2d}

\subsection{Ablation Study}

\subsubsection{Impact of Representation Design}

\begin{figure}[!htbp]
\centering
\includegraphics[width=\linewidth]{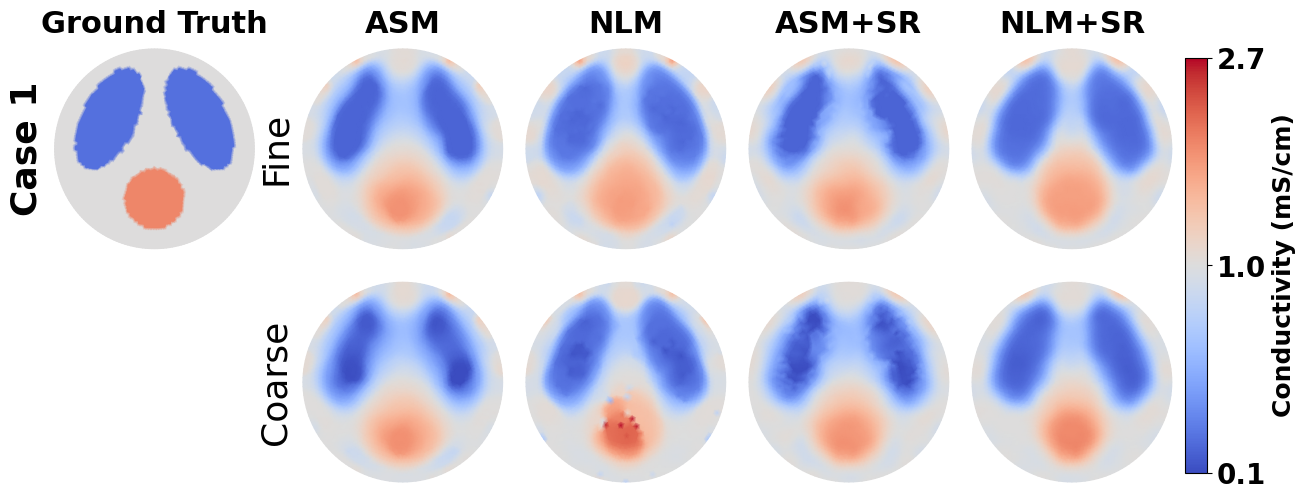}
\caption{Ablation study on ASM and NLM with SR. Rows correspond to fine and coarse bounds.}
\label{fig:ablation_H}
\end{figure}

Fig.~\ref{fig:ablation_H} compares ASM and NLM, each with and without the sparse representation (SR).
For SR variants, a fixed number of $N_b=200$ Laplacian eigenvectors is retained.
With ASM, noticeable artifacts appear between adjacent regions, indicating cross-region leakage induced by the linearized parameterization. In contrast, NLM improves the separation between conductivity regions and yields clearer structural boundaries.
When combined with SR, background fluctuations are further suppressed and the reconstructed shapes become more consistent with the ground truth. In particular, NLM+SR (BC-SR) achieves the most accurate geometry recovery, demonstrating improved structural fidelity under both fine and coarse conductivity bounds.

\subsubsection{Effect of Basis Truncation Level}
\begin{figure}[!htbp]
\centering
\includegraphics[width=.9\linewidth]{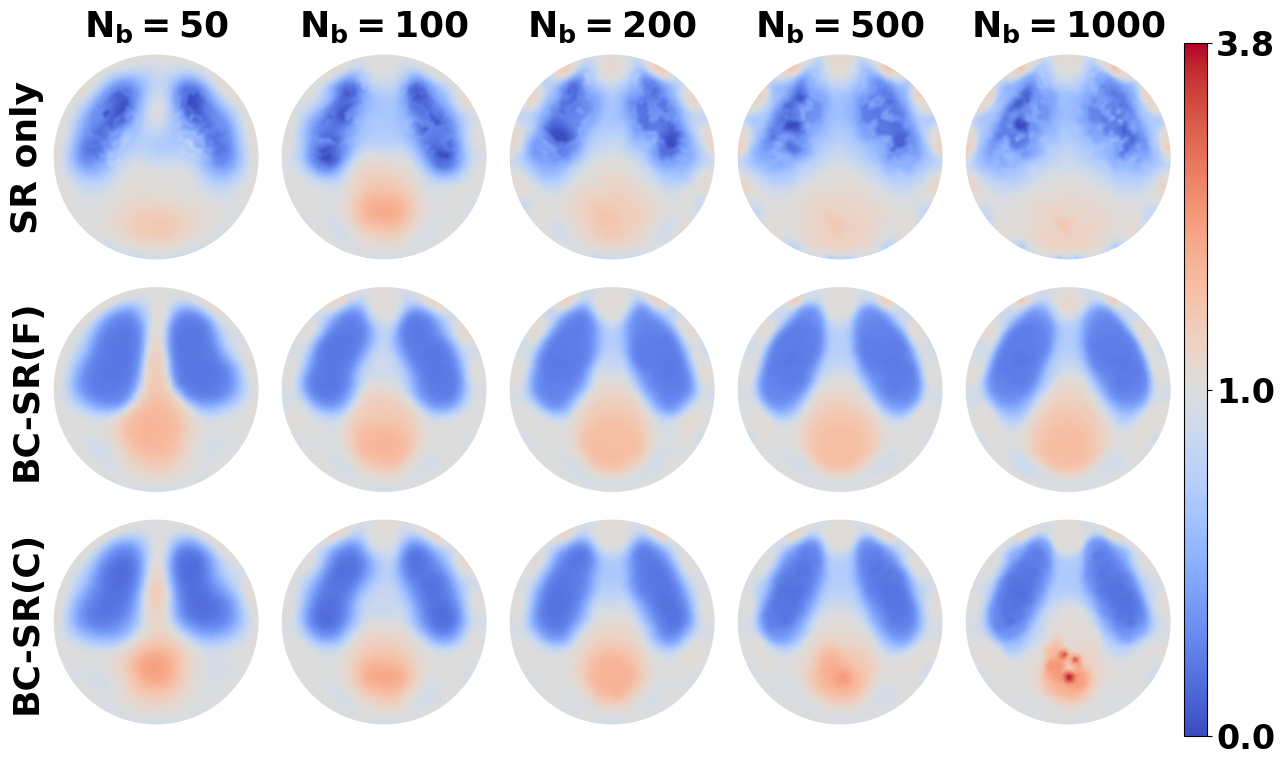}
\caption{Effect of Laplacian basis truncation level $N_b$ on reconstruction performance.}
\label{fig:ablation_K}
\end{figure}

We further investigate the influence of the truncation level $N_b$ in the graph-Laplacian basis. The full conductivity dimension is $N=2038$, and $N_b$ is selected as a fraction of $N$, corresponding to $N_b \in \{50,100,200,500,1000\}$.
As shown in Fig.~\ref{fig:ablation_K}, the reconstruction quality is relatively insensitive to $N_b$ over a wide range, especially when physically meaningful conductivity bounds are imposed. This indicates that the proposed representation is robust to moderate changes in model complexity.
Based on this observation, we set $N_b=0.05N$ for the tank experiments and $N_b=0.1N$ for the 2D and 3D studies, providing a stable trade-off between accuracy and computational cost without case-specific tuning.

\subsection{2D Numerical Phantoms}

\begin{figure}[!htbp]
\centering
\includegraphics[width=0.5\textwidth]{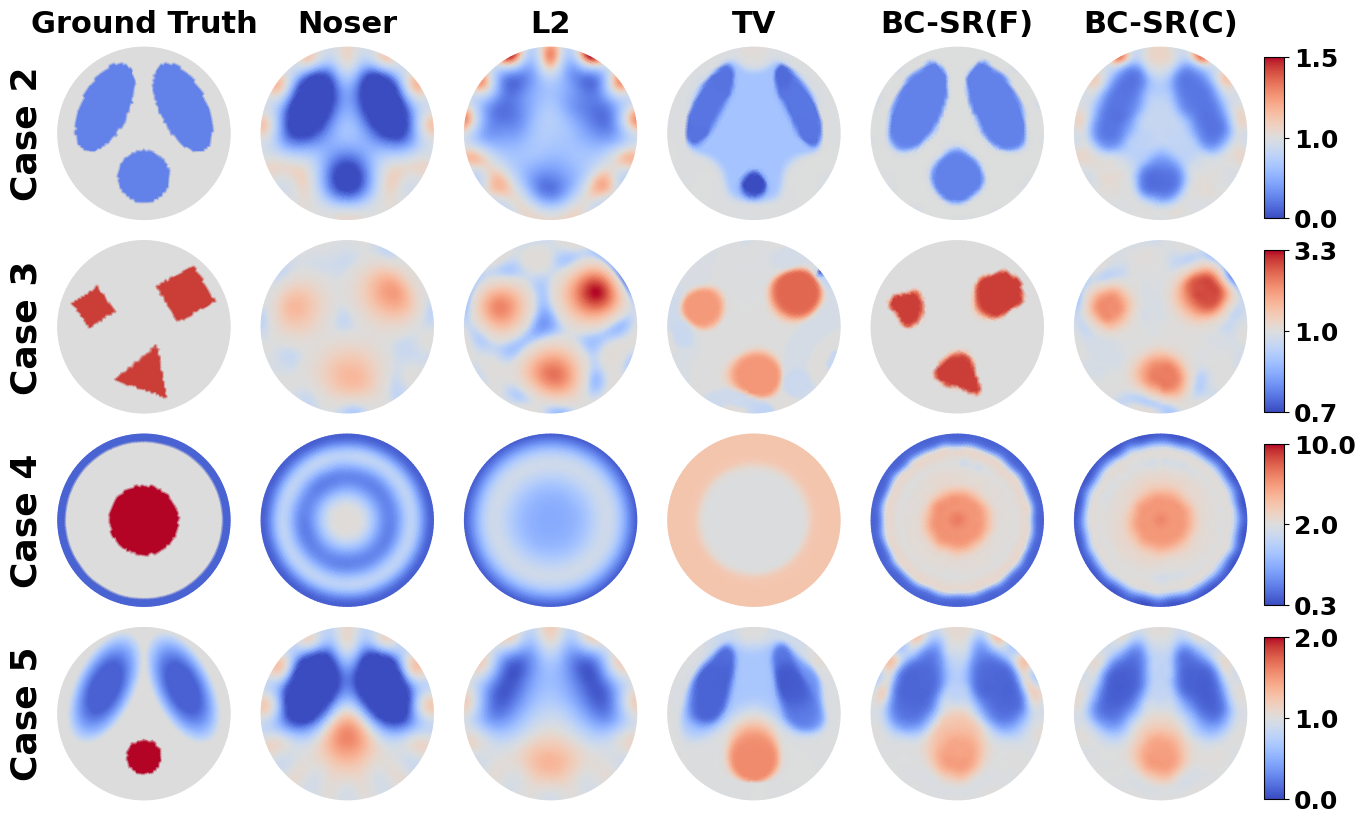}
\caption{Reconstruction results for the 2D numerical phantoms (Cases~2--5) at
60~dB noise. Each row corresponds to one phantom (Cases~2--5 from top to bottom).
Columns show the ground truth (GT), Noser, L2, TV, BC-SR (F), and BC-SR (C).}
\label{fig:2d_cases}
\end{figure}

Fig.~\ref{fig:2d_cases} summarizes the reconstruction results for the 2D numerical phantoms.
NOSER provides a coarse estimate of the conductivity distribution. L2 regularization produces smoother results but blurs sharp boundaries. TV improves edge preservation but may introduce staircase artifacts and fails in low-sensitivity regions, particularly in Case~4.
In contrast, BC-SR achieves consistently better performance across all cases. It preserves sharp interfaces while maintaining smooth variations, successfully reconstructs the shielded structure in Case~4, and remains effective for smoothly varying distributions without relying on a piecewise-constant assumption.
Table~\ref{tab:quantitative} reports SSIM, CC, and RMSE,
providing quantitative support for the consistent performance of BC-SR.

\begin{table}[!ht]
\centering
\caption{Quantitative metrics of reconstruction results for Cases~2--5}
\label{tab:quantitative}
\begin{tabular}{c|c|ccccc}
\toprule
Case & Metric & Noser & L2 & TV & BC-SR(F) & BC-SR(C) \\
\midrule
\multirow{3}{*}{2}
& SSIM & 0.470 & 0.513 & 0.656 & \multicolumn{1}{c}{\textbf{0.896}} & 0.666 \\
& CC   & 0.891 & 0.902 & 0.897 & \multicolumn{1}{c}{\textbf{0.989}} & 0.954 \\
& RMSE & 0.111 & 0.072 & 0.074 & \multicolumn{1}{c}{\textbf{0.009}} & 0.035 \\
\midrule
\multirow{3}{*}{3}
& SSIM & 0.756 & 0.703 & 0.817 & \multicolumn{1}{c}{\textbf{0.879}} & 0.785 \\
& CC   & 0.854 & 0.916 & 0.927 & \multicolumn{1}{c}{\textbf{0.950}} & 0.933 \\
& RMSE & 0.138 & 0.071 & 0.060 & \multicolumn{1}{c}{\textbf{0.039}} & 0.059 \\
\midrule
\multirow{3}{*}{4}
& SSIM & 0.655 & 0.708 & 0.460 & \multicolumn{1}{c}{0.724} & \textbf{0.776} \\
& CC   & 0.627 & 0.379 & 0.055 & \multicolumn{1}{c}{0.911} & \textbf{0.919} \\
& RMSE & 0.633 & 0.661 & 0.737 & \multicolumn{1}{c}{\textbf{0.197}} & 0.202 \\
\midrule
\multirow{3}{*}{5}
& SSIM & 0.632 & 0.835 & 0.804 & \multicolumn{1}{c}{0.864} & \textbf{0.881} \\
& CC   & 0.904 & 0.950 & 0.955 & \multicolumn{1}{c}{0.961} & \textbf{0.970} \\
& RMSE & 0.089 & 0.041 & 0.034 & \multicolumn{1}{c}{0.029} & \textbf{0.023} \\
\bottomrule
\end{tabular}
\end{table}

\subsection{Noise-Robustness Study}
\begin{figure}[!htbp]
\centering
\includegraphics[width=\linewidth]{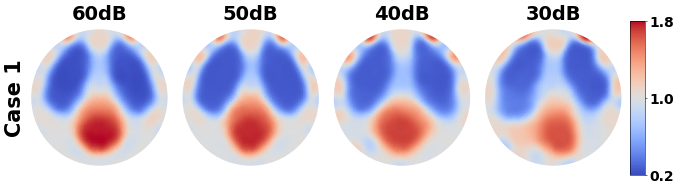}
\caption{Noise-robustness study on Case~1 using BC-SR with the SNR-adaptive TV
weighting in \eqref{eq:alpha_snr}. Columns correspond to different SNR levels
(60, 50, 40, and 30~dB).}
\label{fig:noise_study}
\end{figure}

Fig.~\ref{fig:noise_study} reports Case~1 reconstructions at different noise
levels using \eqref{eq:alpha_snr}, and Table~\ref{tab:noise_metrics} summarizes
SSIM, CC, and RMSE. As SNR decreases from 60~dB to 30~dB,
image quality degrades smoothly and monotonically without abrupt failure,
indicating robust behavior of BC-SR under measurement noise with the proposed
SNR-driven TV weighting.

\begin{table}[!htbp]
\centering
\caption{Quantitative metrics for the noise-robustness study on Case~1 using
BC-SR with SNR-adaptive TV weighting.}
\label{tab:noise_metrics}
\begin{tabular}{c|ccc}
\hline
SNR (dB) & SSIM & CC & RMSE \\
\hline
60 & \textbf{0.737} & \textbf{0.961} & \textbf{0.031} \\
50 & 0.733 & 0.951 & 0.039 \\
40 & 0.694 & 0.940 & 0.047 \\
30 & 0.690 & 0.936 & 0.052 \\
\hline
\end{tabular}
\end{table}

\subsection{Tank Experiments}
\begin{figure}[!htbp]
\centering
\includegraphics[width=\linewidth]{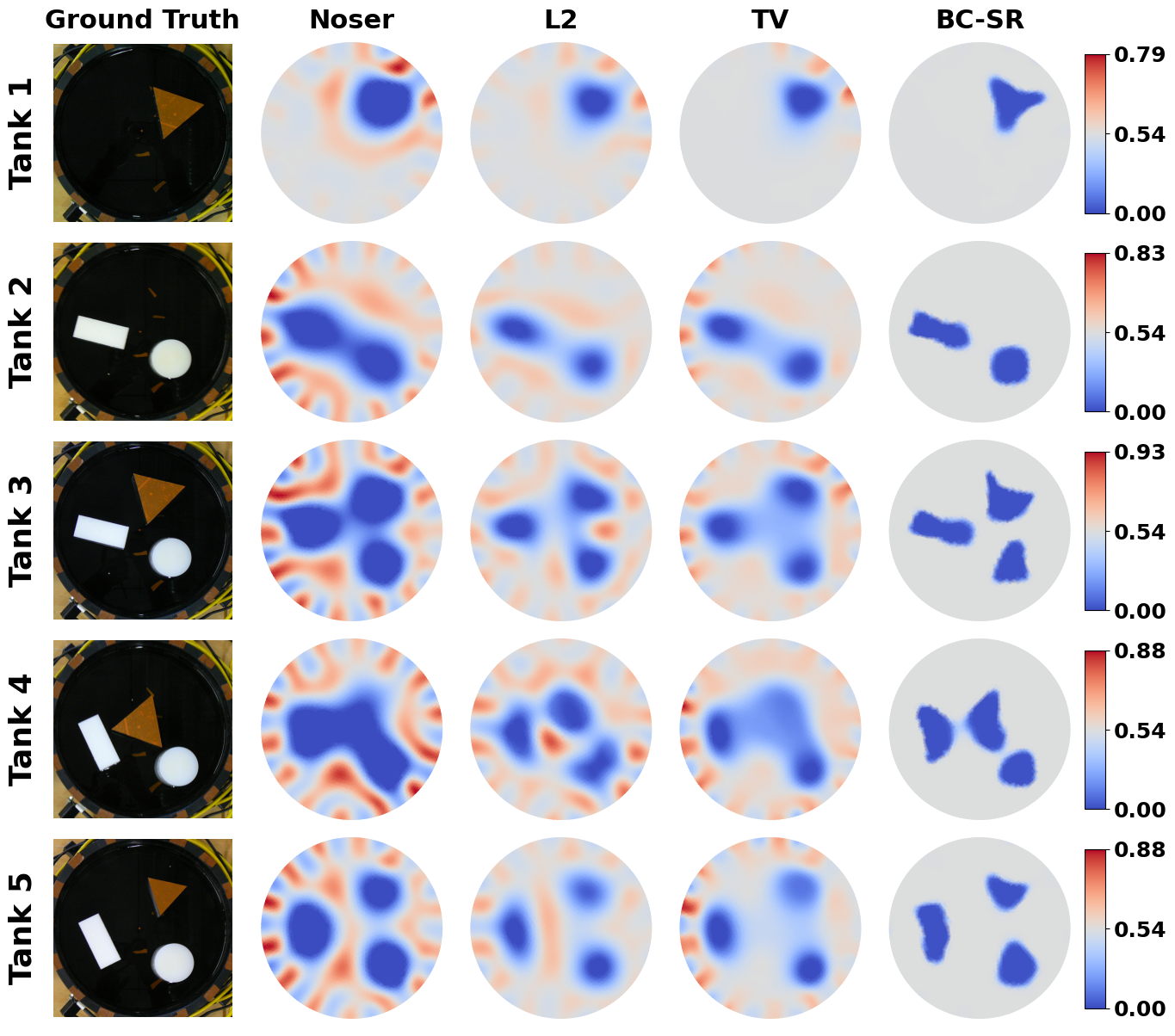}
\caption{Reconstruction results for five tank experiments with saline background
and insulating agar inclusions. Each row corresponds to one tank configuration
(Tank~1--Tank~5), and columns show reconstructions obtained by Noser, L2, TV, and
BC-SR.}
\label{fig:tank_results}
\end{figure}

Fig.~\ref{fig:tank_results} compares reconstructions for five different tank experiments. NOSER is fast but produces blurred boundaries; L2 tends to oversmooth and
merge nearby inclusions; TV preserves edges but may introduce staircase artifacts
and shape distortions for irregular objects. BC-SR yields cleaner backgrounds and
more accurate inclusion number and shape across all tank configurations.

\section{3D Results}
\label{sec:results_3d}

\subsection{3D Numerical Phantoms}

\subsubsection{Case~2.5D}
\begin{figure}[!htbp]
    \centering
    \includegraphics[width=\linewidth]{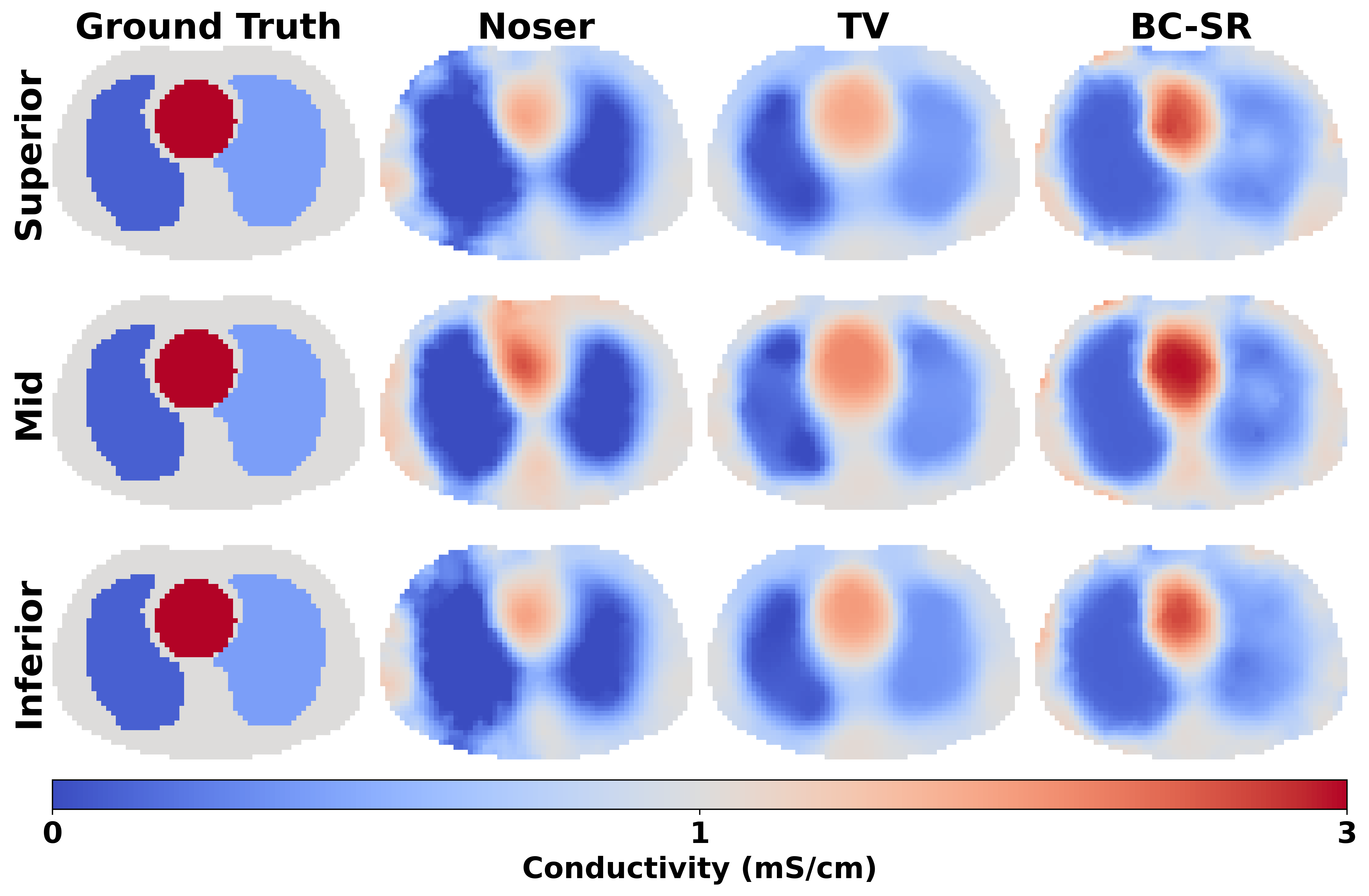}
    \caption{3D simulation results for Case~2.5D.
    From top to bottom, three axial slices at normalized positions
    $z=0.4$, $z=0.0$, and $z=-0.4$ are shown for the ground truth
    and the reconstructions obtained by Noser, TV, and BC-SR.}
    \label{fig:3d_case1}
\end{figure}

We first consider the heart--lung phantom in Fig.~\ref{fig:3d_phantoms}(a), which is homogeneous along the $z$-direction. Each axial slice contains fixed left-lung, right-lung, and heart regions with prescribed conductivities (Table~\ref{tab:phantom_params}). Representative axial slices are shown in Fig.~\ref{fig:3d_case1}.

NOSER captures the overall organ structure but exhibits limited spatial resolution and amplitude bias. TV improves structural delineation but tends to shrink lung regions and underestimate the heart conductivity. In comparison, BC-SR achieves more accurate shape and intensity reconstruction, with only minor residual non-uniformity in the right lung.

\subsubsection{Case~3D}
\begin{figure}[!htbp]
    \centering
    \includegraphics[width=\linewidth]{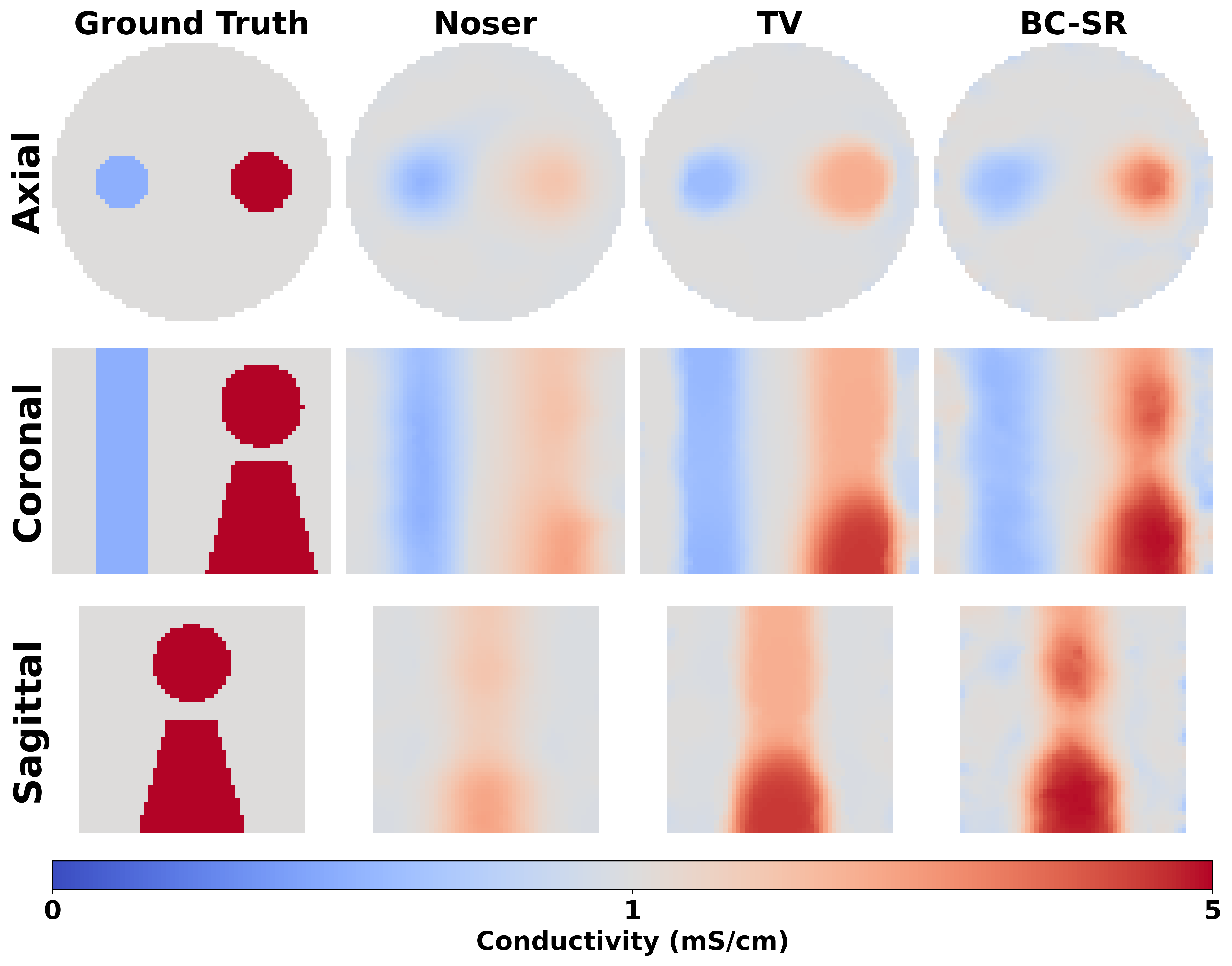}
    \caption{3D simulation results for Case~3D.
    An axial slice at $z=0.2$, a coronal slice at $y=0$, and a sagittal slice
    at $x=0.6$ are shown for the ground truth and the reconstructions
    obtained by Noser, TV, and BC-SR.}
    \label{fig:3d_case2}
\end{figure}

We next evaluate a fully 3D phantom with strong $z$-variation (Fig.~\ref{fig:3d_phantoms}(b)), consisting of a low-conductivity cylinder and two high-conductivity inclusions (a sphere and a frustum) embedded in a homogeneous background. Representative axial, coronal, and sagittal slices are shown in Fig.~\ref{fig:3d_case2}.

NOSER fails to recover the full 3D structure, particularly underestimating high-conductivity regions. TV partially restores the lower inclusion but struggles with the upper sphere. BC-SR consistently reconstructs both inclusions with more accurate geometry and conductivity levels, achieving the best overall balance between structural fidelity and quantitative accuracy.

\begin{figure*}[!ht]
    \centering
    \includegraphics[width=.9\textwidth]{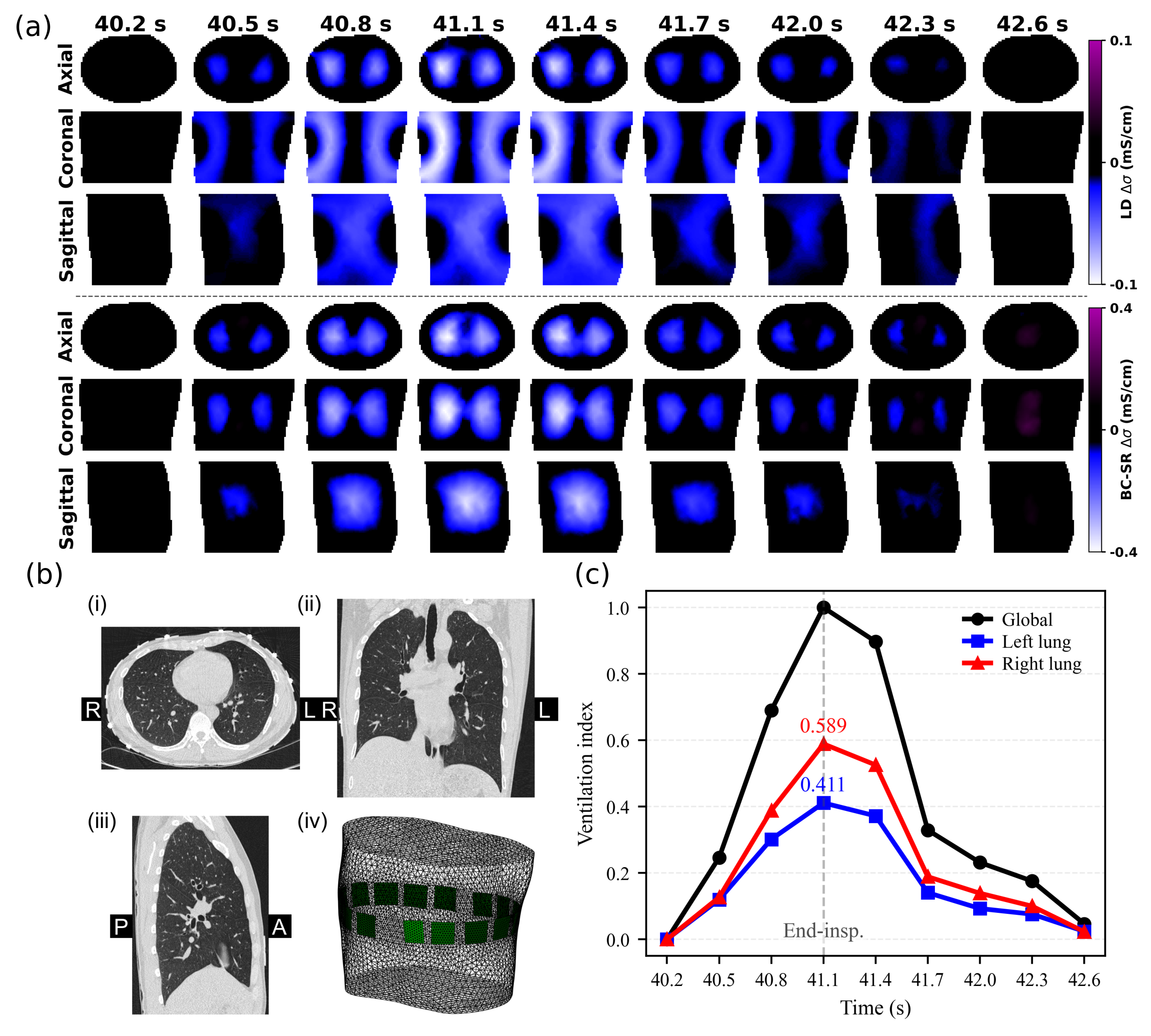}
   \caption{In-vivo  time-difference imaging results over a full breathing cycle. 
(a) Comparison between linear difference (LD) with NOSER-type regularization and the proposed BC-SR method. BC-SR provides a 3D conductivity reconstruction, visualized by axial ($z=0$), coronal ($y=0$), and sagittal ($x=-0.3$) slices. 
(b) Corresponding CT slices ((i)--(iii)) for anatomical reference, along with the FEM mesh (iv). 
(c) Temporal evolution of the lung ventilation index $F_\Omega$ over the respiratory cycle.}
    \label{fig:clinical}
\end{figure*}

\subsection{In-vivo Experiments}

Fig.~\ref{fig:clinical} summarizes the results from three complementary perspectives: (a) spatial reconstruction, (b) anatomical reference, and (c) temporal ventilation analysis.
Fig.~\ref{fig:clinical}(a) compares the reconstructed 3D conductivity distributions obtained by LD and BC-SR. LD provides a qualitatively interpretable reconstruction but exhibits strong smoothing effects and limited spatial consistency across slices. In contrast, BC-SR yields more coherent and spatially consistent 3D structures across axial ($z=0$), coronal ($y=0$), and sagittal ($x=-0.3$) views, enabled by its global 3D basis representation with bound-constrained parameterization, which inherently promotes structured and spatially coherent volumetric distributions beyond slice-wise or plane-dominated reconstructions.
Due to the ill-posed nature of EIT, regions away from the electrode plane are inherently weakly constrained by measurements. In this context, the reconstruction is governed by the global low-frequency 3D representation, and the resulting off-plane structures should be interpreted as physics-consistent volumetric inferences induced by the model, rather than direct observations or reconstruction artifacts.
Fig.~\ref{fig:clinical}(b) presents CT slices ((i)--(iii)) as anatomical references for qualitative assessment of lung morphology, since no ground-truth conductivity is available in in-vivo measurements. The FEM mesh used for EIT modeling is also shown in (iv). BC-SR demonstrates better agreement with the expected lung anatomy at end-inspiration ($41.1~\mathrm{s}$), while LD exhibits noticeable geometric distortions, including overextension of high-conductivity regions toward the boundary.
To further analyze functional ventilation dynamics, we define a lung ventilation index $F_\Omega$ over a region $\Omega$ by integrating conductivity decreases during the respiratory cycle:
\begin{equation}
F_\Omega
=
\sum_{\Gamma\in\Omega}\bigl[-\overline{\Delta\sigma}_\Gamma\bigr]_+\,\mathcal{V}_\Gamma,
\label{eq:vent_index}
\end{equation}
Let $\mathcal{N}_\Gamma$ denote the set of nodes associated with element $\Gamma$, and define $N_\Gamma = |\mathcal{N}_\Gamma|$ as the number of nodes. The element-wise mean conductivity change is then given by
\begin{equation}
\overline{\Delta\sigma}_\Gamma
=
\frac{1}{N_\Gamma}
\sum_{j \in \mathcal{N}_\Gamma}
\Delta\sigma_j.
\end{equation}

Fig.~\ref{fig:clinical}(c) shows the temporal evolution of the ventilation index $F_\Omega$ over a full breathing cycle. Ventilation increases in both lungs during inspiration, with a faster rise in the right lung. The peak occurs at $41.1~\mathrm{s}$, with left and right contributions of $41.1\%$ and $58.9\%$, respectively. The inspiration-to-expiration duration ratio is approximately $1:1.7$.

This left--right distribution is physiologically consistent with lung anatomy: the right lung (three lobes) generally has greater volume and ventilation capacity than the left (two lobes) \cite{west2020west}, with typical volume ratios of $40\%$--$45\%$ (left) and $55\%$--$60\%$ (right) \cite{nunn2013applied}. The BC-SR results fall within this range, supporting their quantitative validity.
In contrast, LD relies on local linearization and shows reduced temporal stability, limiting its suitability for volumetric analysis. Consequently, the proposed 3D ventilation index is not evaluated for LD.
\section{Discussion}
\label{discussion}
The proposed BC-SR framework integrates structural priors and physical constraints through a nonlinear parameterization, enabling improved stability and reconstruction quality across a range of EIT scenarios. The in-vivo results further demonstrate its potential for quantitative 3D difference imaging under realistic measurement conditions.
A key advantage of the proposed formulation lies in its representation-driven design. By restricting the solution space through a low-dimensional basis and enforcing admissible conductivity ranges via a nonlinear mapping, BC-SR improves conditioning and reduces the sensitivity to noise and modeling errors. In particular, the warm-start strategy aligns the reconstruction with the baseline state, thereby reducing shared modeling errors and improving robustness in time-difference imaging.

Despite these advantages, several aspects merit further consideration. First, the reconstruction performance depends on the choice of conductivity bounds, which are assumed to be known or estimated as a prior. While the method is relatively robust to moderate bound variations, overly restrictive or excessively loose bounds may affect reconstruction accuracy.
Another aspect is that the expressiveness of the representation is controlled by the number of retained basis functions $N_b$, which governs the trade-off among reconstruction fidelity, computational efficiency, and robustness. Although the proposed method is stable across a wide range of $N_b$, larger values increase representation capacity but may lead to overfitting, particularly in noisy settings.
The current framework also relies on a fixed graph-based basis derived from mesh connectivity. While this provides a generic and geometry-consistent prior, it does not explicitly adapt to problem-specific features or anatomical structures. Incorporating adaptive or data-driven basis constructions could further enhance representation capability.
Lastly, the computational cost of the nonlinear optimization is higher than that of linearized methods such as the LD baseline, which may limit real-time applications. However, the improved reconstruction accuracy and robustness may justify this trade-off in applications requiring quantitative analysis.
Future work will focus on adaptive basis design, improved initialization strategies, and efficient implementations for real-time clinical deployment.

\section{Conclusion}
\label{conclusion}
We proposed BC-SR, a novel EIT reconstruction framework based on Bound-Constrained Sparse Representation. By integrating structural priors and physical constraints through a bound-preserving mapping and a truncated graph-Laplacian basis, BC-SR eliminates the need for explicit regularization. This formulation ensures stable and robust reconstructions even in the presence of noise and modeling mismatches, and supports both absolute and time-difference imaging in a unified framework.
Extensive validation through 2D/3D simulations, tank experiments, and in-vivo data demonstrates that BC-SR significantly improves structural fidelity and quantitative consistency compared to conventional methods. These results highlight BC-SR’s potential for reliable, robust ventilation-related analysis, particularly in clinical applications.
The flexibility of the BC-SR framework, combined with its physics-consistent approach, makes it a promising tool for addressing other nonlinear inverse problems with boundary measurements.
\section*{Acknowledgment}
The authors would like to sincerely thank Professor Maokun Li from the
Department of Electronic Engineering, Tsinghua University, for providing the
in-vivo data used in this study. The authors also thank Junwu Wang for
providing the 3D forward-solver code and for helpful discussions,
and Bowen Tong for valuable comments and suggestions.

\bibliographystyle{IEEEtran}
\bibliography{reference}
\end{document}